\documentclass{article}

\usepackage{microtype}
\usepackage{graphicx}
\usepackage{subcaption}
\usepackage{booktabs} % for professional tables
\usepackage{multirow} % 必须宏包：合并行
\usepackage{amsmath}  % 必须宏包：数学符号
\usepackage{hyperref}

\usepackage[preprint]{icml2026}

\usepackage{amsmath}
\usepackage{amssymb}
\usepackage{mathtools}
\usepackage{amsthm}
\usepackage[capitalize,noabbrev]{cleveref}

\theoremstyle{plain}
\newtheorem{theorem}{Theorem}[section]

\theoremstyle{definition}
\newtheorem{definition}[theorem]{Definition}
\newtheorem{assumption}[theorem]{Assumption}
\theoremstyle{remark}

\renewcommand{\algorithmiccomment}[1]{\hfill #1}

\usepackage[textsize=tiny]{todonotes}

\icmltitlerunning{Disentangled Instrumental Variables for Causal Inference with Networked Observational Data}

\begin{document}

\twocolumn[
  \icmltitle{Disentangled Instrumental Variables for Causal Inference with Networked Observational Data}



  \icmlsetsymbol{equal}{*}

  \begin{icmlauthorlist}
    \icmlauthor{Zhirong Huang}{gxnu}
    \icmlauthor{Debo Cheng}{HNU}
    \icmlauthor{Guixian Zhang}{MTU}
    \icmlauthor{Yi Wang}{gxnu}
    \icmlauthor{Jiuyong Li}{SAU}
    \icmlauthor{Shichao Zhang}{gxnu}
  \end{icmlauthorlist}

  \icmlaffiliation{gxnu}{School of Computer Science and Engineering; Guangxi Key Lab of MSMI; MOE Key Lab of EBIT, Guangxi Normal University, Guilin, 541004, Guangxi, China}
  \icmlaffiliation{HNU}{School of Computer Science and Technology, Hainan University, Haikou, Hainan 570228, China}
    \icmlaffiliation{SAU}{School of Information Technology and Mathematical Sciences, University of South Australia, Adelaide, SA 5095, Australia.}
    \icmlaffiliation{MTU}{School of Computer Science and Technology,China University of Mining and Technology,Xuzhou,221116,Jiangsu,China}

  \icmlcorrespondingauthor{Debo Cheng}{chengd@hainanu.edu.cn}

  \icmlkeywords{Machine Learning, ICML}

  \vskip 0.3in
]



\printAffiliationsAndNotice{}  % no special notice (required even if empty)

\begin{abstract}
  Instrumental variables (IVs) are crucial for addressing unobservable confounders, yet their stringent exogeneity assumptions pose significant challenges in networked data. Existing methods typically rely on modelling neighbour information when recovering IVs, thereby inevitably mixing shared environment-induced endogenous correlations and individual-specific exogenous variation, leading the resulting IVs to inherit dependence on unobserved confounders and to violate exogeneity. To overcome this challenge, we propose \underline{Dis}entangled \underline{I}nstrumental \underline{V}ariables (DisIV) framework, a novel method for causal inference based on networked observational data with latent confounders. DisIV exploits network homogeneity as an inductive bias and employs a structural disentanglement mechanism to extract individual-specific components that serve as latent IVs. The causal validity of the extracted IVs is constrained through explicit orthogonality and exclusion conditions. Extensive semi-synthetic experiments on real-world datasets demonstrate that DisIV consistently outperforms state-of-the-art baselines in causal effect estimation under network-induced confounding.
\end{abstract}

% Instrumental Variables (IVs) are widely used to address unobserved confounding. Existing methods are constrained by the independent and identically distributed assumption or the coarse utilisation of neighbour information, making it difficult to effectively separate individual-specific variation from shared confounding factors within the entangled features of networked data. This consequently suppresses latent exogenous signals.

% Instrumental variables (IVs) are crucial for addressing unobservable confounding, yet their stringent exogeneity assumptions face significant challenges in network data. Existing methods typically rely on modelling neighbourhood information when recovering instrumental variables, thereby encoding both shared environment-induced endogenous correlations and individual-level exogenous variation into the IV representation, undermining their exogeneity.
  
  % , yet their construction remains challenging in networked observational data due to feature entanglement. Existing methods often fail to distinguish between individual-specific variation and shared confounding factors, thereby suppressing latent exogenous signals. 

\section{Introduction}

    Accurately estimating Individual Treatment Effects (ITE) is fundamental to the transition from relationship discovery to intelligent decision-making, with broad applications in recommendation systems~\cite{huang2025learning}, precision medicine~\cite{ma2023look}, and the social sciences~\cite{imbens2024causal}. ITE estimation aims to answer a counterfactual question: ``How would the outcome change if an intervention were applied to a specific individual?''. By addressing this question, causal inference provides a principled foundation for individualised decision-making. However, Randomised Controlled Trials (RCTs), the gold standard for causal inference, are often challenging due to high costs and ethical limits. Consequently, causal inference using observational data has emerged as a widely adopted alternative.

Although observational data provides a viable basis for causal inference, many real-world applications involve networked data~\cite{ma2022learning}. In networked observational data, individuals are interconnected rather than independent, and treatment assignments and outcomes depend on both individual and neighbour information~\cite{chen2024doubly,jiang2022estimating}. Such network dependence violates standard assumptions, including the Independent and Identically Distributed (IID) sampling assumption and the Stable Unit Treatment Value Assumption (SUTVA)~\cite{rubin1974estimating}. Furthermore, unobserved confounders correlated with the network structure introduce additional sources of bias, leading to biased causal effect estimation.

The Instrumental Variable (IV) approach is a widely employed method for addressing the effects of unobserved confounders~\cite {pearl2009causality,angrist1996identification,cheng2024instrumental}. IVs constitute a class of exogenous variables that are correlated with the treatment but affect the outcome only through the treatment and are independent of unobserved confounders~\cite{pearl2009causality}. Classical IV methods, such as two-stage least squares (2SLS)~\cite{angrist1995two}, employ two-stage regression to eliminate endogeneity. With the advancement of deep learning~\cite{cai2024granger,zhang2024learning}, Hartford et al.~\cite{hartford2017deep} proposed DeepIV, introducing deep neural networks to model complex non-linear distributions in data. However, a fundamental limitation of existing IV-based methods lies in the challenge of identifying valid IVs. Most approaches depend on expert-specified exogenous variables, whose validity is difficult to justify or empirically assess, particularly in complex observational settings with pervasive unobserved confounding.

% Singh et al. proposed KernelIV, which employs kernel functions to map data into a reproducing kernel Hilbert space (RKHS) for two-stage regression

% However, despite their strong performance on independent samples, these methods fail severely in networked scenarios. First, these methods all rely on the I.I.D. assumption and ignore network structure. Second, they depend on expert-defined strong exogenous variables, yet in intricate networked data, researchers struggle to manually identify a valid IV.

\begin{figure}[t]
  \vskip 0.2in
  \begin{center}
    \centerline{\includegraphics[width=0.6\columnwidth]{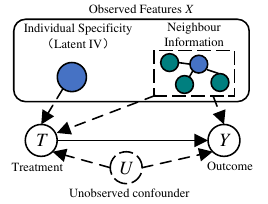}}
    \caption{Conceptual diagram illustrating the discovery of latent IVs in networked data. Within the observed feature $X$, neighbour information is entangled with individual-specific information. Our method identifies the individual-specificity information as a latent IV. By structurally disentangling $X$, this latent IV can be extracted to eliminate bias introduced by unobserved confounder $U$ and neighbour information.
    }
    \label{introduction}
  \end{center}
\end{figure}

Recent research has focused on discovering latent IVs to address the scarcity of predefined IVs. For instance, Li et al.~\cite{li2024distribution} proposed a deep Variational autoencoder based IV method (VIV) for generating IVs via variational inference and enforces exogeneity through adversarial training. In network settings, the Networked Instrumental (NetIV) regression framework proposed by Zhao et al.~\cite{zhao2024networked} constructs IVs using neighbour information to extend classical IV methods. While these approaches demonstrate effectiveness in IV recovery, important limitations remain in networked observational data. In real-world networks, neighbour information often reflects shared environments and homophily~\cite{zhao2024learning}. When neighbour information simultaneously affects both treatment and outcomes, directly constructing IVs from such information risks introducing environmental confounders, thereby violating the exogeneity assumption. Consequently, a central challenge in learning IVs from networked observational data is how to exploit network structure while avoiding the misidentification of environmental confounders as IVs and preserving exogeneity.

To address these challenges, we propose a Disentangled IV Representation Learning (DisIV) framework for estimating causal effects from networked observational data with unobserved confounders. DisIV aims to recover valid latent IVs by capturing individual-specific variation. As shown in Figure~\ref{introduction}, DisIV is grounded in the premise that neighbour information largely characterises the shared environment and structural confounding~\cite{guo2020learning}, whereas individual-specific information preserves independent exogenous signals, rendering it suitable as a latent IV~\cite{heckman2006understanding}. Following this premise, DisIV explicitly decomposes the generative mechanism of individual representations at the latent level, separating environmental components induced by network structure from an individual’s own exogenous variation. This decomposition enables the construction of latent IV representations. To guarantee exogeneity, DisIV imposes explicit structural disentanglement constraints that enforce statistical orthogonality between the latent IVs and network-induced environmental factors. Moreover, DisIV strictly enforces the exclusion restriction by explicitly modelling environmental influences on outcomes, thereby ensuring that the latent IV affects the outcome exclusively through the treatment pathway. In summary, our principal contributions are as follows:
\begin{itemize}
    \item We propose a causal perspective for characterising valid IVs as individual-specific latent factors, providing new insights into IV modelling in networked settings.
    \item We propose the Disentangled Instrumental Variables (DisIV) framework that recovers latent IVs by structurally disentangling individual exogenous variations from network-induced environmental confounding.
    \item We conduct extensive experiments on two semi-synthetic datasets constructed from real-world social networks, demonstrating the effectiveness and superiority of our DisIV method.
\end{itemize}

% Building on this, DisIV imposes stringent structural disentangling constraints to ensure statistical orthogonality between the latent IVs and the network environmental information, thereby guaranteeing the exogeneity of the IVs.

% Furthermore, the framework strictly adheres to the exclusion restriction by explicitly channelling environmental factors to account for outcome variations, thereby isolating the IV's influence solely through the treatment path. 

% Specifically, this study is modeled on the basis of structural dependencies induced by network homogeneity. In network systems, neighbourhood information primarily reflects the shared environment and structural homogeneity surrounding individuals. Such information frequently influences both treatment assignment and outcome generation, thereby constituting a significant source of structural confounding. In contrast, information related to individual processing decisions but not directly reflecting the network environment is more likely to manifest as idiosyncratic variation at the individual level. Within the causal inference framework, such idiosyncratic variation inherently possesses modelling potential as instrumental variable candidates. 

\section{Related Work}
\subsection{Causal Inference under Ignorability Assumption}
Early research primarily relied on the strong Ignorability assumption, focusing on eliminating distributional differences through balanced representation learning~\cite{shi2019adapting,louizos2017causal,yao2018representation}. For instance, Shalit et al.~\cite{shalit2017estimating} proposed Counterfactual Regression (CFR) and Treatment-Agnostic Representation Network (TARNet), employing minimised Integral Probability Metrics (IPM) to constrain distributional bias between treatment groups. Additionally, tree-based models (such as BART~\cite{chipman2010bart} and Causal Forest~\cite{athey2019estimating}) are frequently employed to address non-linear confounding. However, these methods face significant challenges when applied to networked data: they typically assume samples are IID, overlooking dependencies induced by network structure; more critically, they cannot address unobserved confounders, leading to biased estimation when latent biases exist.
% Shi et al. designed Dragonnet, which achieves end-to-end correction of causal effect bias by introducing a propensity score prediction head combined with Targeted Minimum Loss Estimation (TMLE)-based objective regularisation.

\subsection{Causal Effect Estimation on Networked Data}
Given the ubiquity of network data, recent research has employed Graph Convolutional Networks (GCNs)~\cite{kipf2016semi} to capture confounding patterns in networks. Representative work, such as Guo et al.'s Network Deconfounder (NetDconf)~\cite{guo2020learning}, utilises GCNs to aggregate neighbour information as a proxy for unobserved confounders. Subsequent studies have further incorporated techniques such as geometric representation learning to enhance the robustness of confounding estimation~\cite{jiang2022estimating}. For instance, Cui et al.~\cite{cui2024treatment} introduced hyperbolic geometric embeddings to capture intricate higher-order structural relationships. Whilst existing methods utilise network homogeneity to supplement confounding information, they erroneously treat all neighbour representations as confounders, overlooking the latent individual specificity. This over-adjustment diminishes the exogenous variance of treatment variables, thereby compromising the precision of causal effect estimations.

% Whilst these approaches supplement confounding information by leveraging network homogeneity, they typically rest on an implicit strong assumption: all representations learned from network neighbourhoods should be regarded as confounding factors and adjusted for. This strategy overlooks individual-specific variation that may serve as latent IVs. Such over-adjustment deprives the model of exogenous variability in the treatment variable, thereby compromising estimation accuracy.

\subsection{Instrumental Variable Methods for Causal Effect Estimation}
When unobserved confounders are present, IV methods are commonly employed to achieve causal identification~\cite{cheng2023conditional,cheng2023discovering}. With the advancement of deep learning~\cite{huang2025interaction,zhang2025modality}, the paradigm of IV methods has evolved from classical 2SLS~\cite{angrist1995two} to non-linear DeepIV~\cite{hartford2017deep}. To address the scarcity of predefined IVs, recent research has turned to latent IV discovery~\cite{wu2025instrumental}, such as AutoIV~\cite{yuan2022auto}, which aims to automatically decouple latent factors that satisfy the IV definition from high-dimensional observed data. For network data, NetIV~\cite{zhao2024networked} constructs IVs using neighbour information within network topologies, extending the classical IV causal inference framework to network settings. However, existing methods still face significant limitations when handling networked feature entanglement: approaches relying on explicit IVs are constrained by a lack of prior knowledge; traditional paradigms based on the IID assumption, which treats instances as isolated events, neglect network dependencies; and existing IV methods for networked data often lack refined disentangling mechanisms, making it difficult to effectively distinguish environmental assimilation from individual specificity. To address these challenges, DisIV leverages network homogeneity as an inductive bias to automatically recover latent IVs via structural disentangling.

% However, these methods exhibit significant limitations when handling networked feature entanglement: DeepIV relies on expert a priori knowledge that is difficult to obtain; existing latent IV discovery methods are largely constrained by the I.I.D. assumption, neglecting network dependencies; while NetIV utilises topological structure, it lacks a refined disentangling mechanism for intra-node features, making it difficult to effectively distinguish between environmental assimilation features and individual specificity. Addressing these challenges, the proposed DisIV method pioneers the use of network homogeneity as an inductive bias. Through structural disentangling, it automatically recovers latent IVs from entangled features, thereby enabling unbiased causal inference in complex network settings.

\section{Preliminary}
\subsection{Notation}
We formally define networked observational data as an attribute graph $\mathcal{G} = (\mathbf{A}, \mathbf{X})$ comprising $N$ nodes. Here, $\mathbf{A} \in \{0, 1\}^{N \times N}$ denotes the adjacency matrix describing the topological structure of $\mathcal{G}$. If nodes $i$ and $j$ are connected, then $A_{ij}=1$; otherwise, $A_{ij}=0$. $\mathbf{X} \in \mathbb{R}^{N \times D}$ is the feature matrix, where the $i$-th row $\mathbf{x}_i \in \mathbb{R}^D$ represents the feature information of individual $i$. Following established research conventions~\cite{jiang2022estimating,cai2023generalization,zhao2024networked}, we focus on first-order connections between nodes. The neighbour set of individual $i$ is defined as $\mathcal{N}_i = \{j \mid \mathbf{A}_{ij} = 1\}$. Furthermore, for each individual $i$, we consider a binary treatment variable $t_i \in \{0, 1\}$ (where $t_i=1$ denotes acceptance of intervention, and $t_i=0$ denotes refusal) and a continuous outcome variable $y_i \in \mathbb{R}$. Thus, the networked observational dataset can be represented as $\mathcal{D} = \{\mathbf{A}, \mathbf{X}, \mathbf{t}, \mathbf{y}\}$.

\subsection{Assumptions}
In causal effect estimation, for any individual $i$, only the outcome $y_i$ corresponding to the factual treatment $t_i$ is observable, whilst the counterfactual outcome is typically unattainable. This fundamental limitation is particularly acute in networked observational data, as both treatment assignment $t_i$ and outcome $y_i$ may be simultaneously influenced by network-related confounders (denoted as $\mathbf{u}_i$)

To ensure the theoretical feasibility of identifying ITE from observational data, we introduce the following assumptions~\cite{guo2020learning,zhao2024networked,chen2024doubly}:

\begin{assumption}[Network Consistency]
For any individual $i$, the observed outcome $y_i$ corresponds to the potential outcome under the assigned treatment $t_i$:
    \begin{equation}
        y_i = t_i Y_i(1) + (1 - t_i) Y_i(0).
    \end{equation}
Here, $Y_i(1)$ and $Y_i(0)$ denote the potential outcomes for individual under treatment  ($t_i=1$) and control ($t_i=0$), respectively.
\end{assumption}

\begin{assumption}[Network Overlap]
Given an individual's observed covariates $\mathbf{x}_i$ and confounders $\mathbf{u}_i$, the propensity score for treatment strictly lies between 0 and 1: 
    \begin{equation}
        0 < P(t_i = 1 \mid \mathbf{x}_i, \mathbf{u}_i) < 1.
    \end{equation}
\end{assumption}

\begin{assumption}[Structured Generative Mechanism]
We posit that an individual's observed covariates are generated by the interaction of two latent components: $\mathbf{u}_i$ (confounder) and $\mathbf{z}_i$ (specificity factor). Specifically, we assume the existence of a function $g(\cdot)$ such that:
\begin{equation}
    \mathbf{x}_i = g(\mathbf{z}_i, \mathbf{u}_i).
\end{equation}
\end{assumption}
% This hypothesis characterises the latent generative process of networked observational data, wherein individual-specific information pertinent to causal effect identification becomes entangled with network confounding factors within the observational space.

\subsection{Problem Setup}
Based on the aforementioned assumptions, we aim to estimate causal effects from networked observational data. For any individual $i$ within the network, we focus on the ITE, defined as:
\begin{equation}
\tau_i = Y_i(1) - Y_i(0),
\end{equation}

and the average treatment effect (ATE), defined as:
\begin{equation}
\tau_{ATE} = \mathbb{E}[\tau_i].
\end{equation}

However, in the presence of latent network-related confounders, estimating these causal effects directly from observational data typically yields bias without additional assumptions or auxiliary information.

\begin{figure*}[t]
  \vskip 0.2in
  \begin{center}
    \centerline{\includegraphics[width=0.9\textwidth]{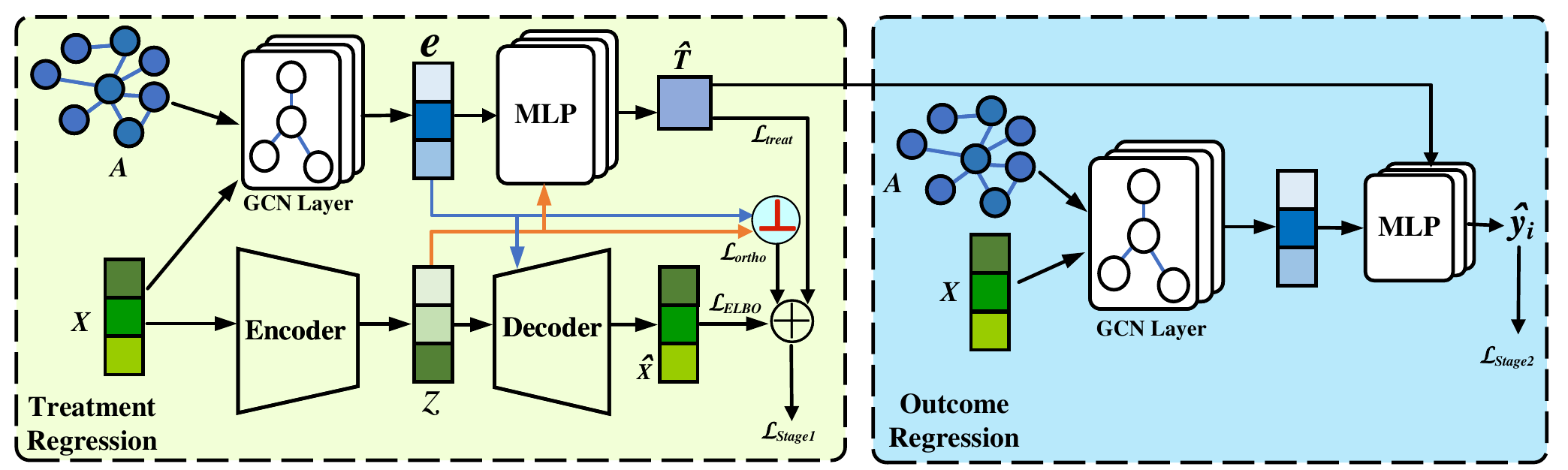}}
    \caption{Overall architecture of DisIV. First, DisIV extracts a confounder proxy $\mathbf{e}$ from neighbour information. Subsequently, during the inference phase, the encoder extracts a latent representation $\mathbf{z}$ as a candidate IV; in the generation phase, we condition on $\mathbf{e}$ and $\mathbf{z}$ to reconstruct the observed data. This asymmetric design compels $\mathbf{z}$ to capture solely residual information (i.e., individual specificity) that cannot be explained by $\mathbf{e}$. Furthermore, the constraint term is introduced to further purify $\mathbf{z}$, ensuring it satisfies the definition of an IV.}
    \label{model}
  \end{center}
\end{figure*}

In real-world networked data, the observed features $\mathbf{x}_i$ are inherently entangled, amalgamating intrinsic attributes with environmental confounding induced by homogeneity. To extract clear causal signals from this complex entanglement, we formulate the networked causal inference problem as a latent variable disentanglement problem. Specifically, to eliminate bias introduced by network confounding, we aim to recover latent individual-specific factors $\mathbf{z}_i$ from entangled observed features $\mathbf{x}_i$ and employ these as IVs for causal effect estimation. 
% This implies that the individual-specific information crucial for causal effect estimation does not exist in isolation but is highly entangled with network confounding factors within the observational space. Given the scarcity of predefined IVs and the heterogeneity of individual environments, relying on a unified global IV is often infeasible.

To ensure the validity of the recovered $\mathbf{z}_i$, it must strictly satisfy the \textit{Relevance}, \textit{Exclusion}, and \textit{Unconfounded}~\cite{pearl2009causality,angrist1996identification}. Detailed definitions of the IVs are presented in the Appendix~\ref{appendix_IV}.

Formally, given a networked observation dataset $\mathcal{D} = \{\mathbf{A}, \mathbf{X}, \mathbf{t}, \mathbf{y}\}$, our objective is to learn a mapping $\Phi: \mathcal{X} \to \mathcal{Z}$ that decouples latent individual specificity $\mathbf{z}_i$ from entangled features $\mathbf{x}_i$. By constraining $\mathbf{z}_i$ to satisfy valid IV conditions, we leverage the recovered factors to mitigate network confounding and achieve unbiased ITE and ATE estimation.

% the core research question addressed herein concerns how to decouple individual-specific factors from $\mathcal{D}$. Specifically, our objective is to learn a disentangling mapping $\Phi: \mathcal{X} \to \mathcal{Z}$ designed to infer the latent specific factor $\hat{\mathbf{z}}_i = \Phi(\mathbf{x}_i)$ from the entangled observed feature $\mathbf{x}_i$, satisfying the causal constraints of an IV in a statistically meaningful manner. Building upon this, we further utilise the recovered IVs $\hat{\mathbf{z}}_i$ to eliminate network confounding bias, thereby ultimately achieving unbiased estimates of both ITE and ATE.

\section{Methodology}
To tackle feature entanglement and IV scarcity in networked data, we propose Disentangled Instrumental Variables (DisIV), a framework that recovers latent IVs from observed data, as shown in Figure~\ref{model}. 

% We derive a variational lower bound for structural disentanglement, augmented by causal regularisation constraints. Implemented via a two-stage training process, DisIV effectively isolates valid instruments to achieve unbiased causal effect estimation.

% To address the challenges of feature entanglement and the absence of effective IVs in causal identification within networked data, we propose the Disentangled IV (DisIV). As illustrated in Figure X, this framework aims to recover IVs from observed data. We first derive an optimisation lower bound for structural disentanglement based on variational inference theory. Subsequently, we introduce causal regularisation constraints, ultimately achieving unbiased causal effect estimation through a two-stage training process.

\subsection{Structural Disentanglement for IV Recovery}
% In real-world scenarios, the information we observe about individuals is typically influenced by neighbourhood environments (confounding factors) and idiosyncrasies. The potentially effective IVs are implicitly embedded within an individual's idiosyncratic information. To decouple this idiosyncratic information from the observed data and thereby recover the IVs, we propose a structural disentangling mechanism, xxx.
Observed features in networked data entangle network-induced environmental confounders with intrinsic individual specificity, the latter of which is the source of valid IVs. To isolate these signals, we propose a structural disentanglement strategy to extract the latent individual specificity required for subsequent IV construction.

To disentangle individual-specific information from the observed features, we first aim to capture the network-related confounders. Under network homogeneity, the local neighbour structure provides informative cues about the unobserved confounder $\mathbf{u}_i$. We employ a GCN as the environment encoder, learning the environment representation $\mathbf{e}_i$ by aggregating v information:
\begin{equation}
\mathbf{e}_i = \text{GCN}(\mathbf{X}, \mathbf{A})_i = \sigma ( \sum_{j \in \mathcal{N}_i \cup \{i\}} \frac{1}{\sqrt{\tilde{d}_i \tilde{d}_j}} \mathbf{W}_{env} \mathbf{x}_j ),
\end{equation}
where $\tilde{d}_i$ denotes the degree of node $i$, and $\mathbf{W}_{env}$ is a learnable weight matrix. The learned $\mathbf{e}_i$ serves as a structural proxy for the confounders.

With the confounder proxy $\mathbf{e}_i$ established, the key challenge is to extract individual-specific information from $\mathbf{x}_i$ that is independent of environmental influences, corresponding to the latent IV $\mathbf{z}_i$. To address this challenge, we employ an asymmetric inference-generation structure to model $\mathbf{z}_i$, as shown in Figure~\ref{model}. In the inference phase, the encoder $q_\phi(\mathbf{z}|\mathbf{x})$ derives the latent variable $\mathbf{z}$ strictly from the observed features. Subsequently, in the generative phase, we introduce the confounder proxy as a condition, utilising the decoder $p_\theta(\mathbf{x}|\mathbf{z}, \mathbf{e})$ to reconstruct the observation. This asymmetric design encourages $\mathbf{z}$ to encode idiosyncratic variations that cannot be explained by the confounder proxy $\mathbf{e}$.

During the inference phase, we employ the Multi-Layer Perceptron (MLP) to map observed features to the distributional space of the latent IVs:
\begin{equation}
    q_\phi(\mathbf{z}_i|x_i) = \mathcal{N}(\mathbf{z}_i|\mu_{\phi}(x_i), \sigma_{\phi}(x_i)),
\end{equation}
where $\mu_{\phi}$ and $\sigma_{\phi}$ are parameterised by the MLP. To enable end-to-end gradient optimisation, we utilise the reparameterisation trick:
\begin{equation}
\mathbf{z}_i = \boldsymbol{\mu}_\phi(\mathbf{x}_i) + \boldsymbol{\sigma}_\phi(\mathbf{x}_i) \odot \boldsymbol{\epsilon}, \quad \boldsymbol{\epsilon} \sim \mathcal{N}(\mathbf{0}, \mathbf{I}),
\end{equation}
where $\mathbf{z}_i$ represents the candidate latent IV that we seek to recover. 

To ensure $\mathbf{z}_i$ accurately captures an individual's intrinsic specificity, the decoder attempts to reconstruct the original features. Crucially, we inject the confounder proxy $\mathbf{e}_i$ as a conditional constraint into the generative process:
\begin{equation} 
p_\theta(\mathbf{x}_i|\mathbf{z}_i, \mathbf{e}_i) = \mathcal{N}(\mathbf{x}_i \mid \boldsymbol{\mu}_{\theta}([\mathbf{z}_i, \mathbf{e}_i]), \mathbf{I}), 
\end{equation}
where $[\cdot , \cdot]$ denotes feature concatenation, and $\boldsymbol{\mu}\theta$ represents the decoder network predicting the reconstruction mean. This architectural design forces $\mathbf{z}_i$ to encode the residual variance in $\mathbf{x}_i$ that cannot be explained by the confounder proxy $\mathbf{e}_i$, thereby structurally ensuring that $\mathbf{z}_i$ constitutes an effective latent IV.

% This architectural design forces $\mathbf{z}_i$ to encode the residual variance in $\mathbf{x}_i$ that cannot be explained by the environment $\mathbf{e}_i$, thereby structurally guaranteeing that $\mathbf{z}_i$ constitutes an effective latent IV.

Based on the asymmetric inference-generation architecture described above, our objective is to maximise the marginal log-likelihood of the observed features $\mathbf{x}$ conditioned on the confounder proxy $\mathbf{e}$:
\begin{equation}
    \log p_\theta(\mathbf{x}_i | \mathbf{e}_i) = \log \int p_\theta(\mathbf{x}_i, \mathbf{z}_i | \mathbf{e}_i) \, d\mathbf{z}_i,
\end{equation}

Since the true posterior of the latent variable is intractable, we introduce the variational posterior $q_\phi(\mathbf{z}_i|x_i)$ and apply Jensen’s inequality to yield the following Evidence Lower Bound (ELBO)~\cite{kingma2013auto}:
\begin{equation}
    \begin{aligned}
        &\log p_\theta(\mathbf{x}_i  | \mathbf{e}_i)
        \\&= \log \int p_\theta(\mathbf{x}_i, \mathbf{z}_i | \mathbf{e}_i) d\mathbf{z}_i 
        \\ &= \log \mathbb{E}_{q_\phi(\mathbf{z}_i|\mathbf{x}i)} \left[ \frac{p_\theta(\mathbf{x}_i, \mathbf{z}_i | \mathbf{e}_i}{q_\phi(\mathbf{z}_i|\mathbf{x}_i)} \right] 
        \\ &\geq \mathbb{E}_{q_\phi(\mathbf{z}_i|\mathbf{x}_i)} \left[ \log p_\theta(\mathbf{x}_i, \mathbf{z}_i | \mathbf{e}_i) - \log q_\phi(\mathbf{z}_i|\mathbf{x}_i) \right]. 
    \end{aligned} 
\end{equation}

We decompose the joint distribution of Eq.(10) into the decoder and the prior distribution:
\begin{equation}
    p_\theta(\mathbf{x}_i, \mathbf{z}_i | \mathbf{e}_i) = p_\theta(\mathbf{x}_i | \mathbf{z}_i, \mathbf{e}_i) p(\mathbf{z}_i | \mathbf{e}_i),
\end{equation}
where we assume that the prior distribution $p(\mathbf{z}_i)$ of $\mathbf{z}$ is independent of $\mathbf{e}$, i.e., $p(\mathbf{z}_i|\mathbf{e}_i) = p(\mathbf{z}_i)$. Furthermore, in line with the literature on variational inference~\cite{kingma2013auto,sohn2015learning,solera2024beta}, the prior is typically assumed to be a Gaussian distribution.

Substituting Eq. (12) into Eq. (11), we obtain the decomposed ELBO:
\begin{equation}
\begin{aligned}
    \mathcal{L}_{\text{ELBO}} =& \mathbb{E}_{q_\phi(\mathbf{z}_i|\mathbf{x}_i)} [\log p_\theta(\mathbf{x}_i | \mathbf{z}_i, \mathbf{e}_i)] \\&- D_{\text{KL}}(q_\phi(\mathbf{z}_i|\mathbf{x}_i) \parallel p(\mathbf{z}_i)), 
\end{aligned}
\end{equation}

where the first term represents the reconstruction loss, while the second term denotes the Kullback-Leibler (KL) divergence. The separability between the latent IV $\mathbf{z}_i$ and the confounder proxy $\mathbf{e}_i$ is induced by the structure of the ELBO. Specifically, the decoder relies on the given $\mathbf{e}_i$ to reconstruct the variation in $\mathbf{x}_i$ associated with the confounding variable. Consequently, the latent IV $\mathbf{z}_i$ is compelled to encode only the residual information required for reconstruction (i.e., individual specificity). Furthermore, the KL divergence acts as an information constraint, imposing a penalty on $\mathbf{z}_i$. Should $\mathbf{z}_i$ encode information related to $\mathbf{e}_i$, this incurs a KL cost, preventing minimisation of the ELBO. The global optimum of the $\mathcal{L}_{\text{ELBO}}$ necessitates that $\mathbf{z}_i$ be statistically independent of $\mathbf{e}_i$, thereby satisfying the exclusion constraint of a valid IV.

\subsection{Optimisation with IV Validity Constraints}
To further ensure the validity of the latent IVs $\mathbf{z}_i$, we introduce regularisation constraints during the optimisation process to strictly adhere to the independence assumption.

First, to satisfy the ``Unconfoundedness'' condition $\mathbf{z}_i \perp \mathbf{u}_i$, we must sever any statistical dependence between the latent IV and the confounder proxy. To this end, we introduce an orthogonality regularisation term that minimises the cosine similarity between $\mathbf{z}_i$ and $\mathbf{e}_i$:
\begin{equation}
    \mathcal{L}_{ortho} = \frac{1}{N} \sum_{i=1}^N \left| \frac{\mathbf{z}_i \cdot \text{sg}(\mathbf{e}_i)}{\| \mathbf{z}_i \|_2 \| \text{sg}(\mathbf{e}_i) \|_2} \right|,
\end{equation}
Crucially, we employ stop-gradient (sg) operations on $\mathbf{e}_i$. This design ensures that the orthogonality constraint acts solely to remove confounder-related information from the latent IV representation, while leaving the confounder proxy itself unaffected and semantically intact.

Secondly, to guarantee IVs correlate with the treatment variable, satisfying the condition $\mathbf{z}_i \not\perp t_i$, latent IVs must encode information about the treatment assignment mechanism. We introduce an auxiliary treatment prediction network $f_T$, whose input is the concatenated latent factors:
\begin{equation}
\mathcal{L}_{treat} = - \frac{1}{N} \sum_{i=1}^N \left[ t_i \log(\hat{t}_i) + (1-t_i) \log(1-\hat{t}_i) \right],
\end{equation}
where $\hat{t}_i = \sigma(f_T([\mathbf{z}_i \| \mathbf{e}_i]))$. This objective serves a dual purpose: it forces $\mathbf{z}_i$ to capture the latent motivation driving the individual to accept treatment and, functionally, serves as the treatment regression in the standard IV framework.

Finally, to satisfy the ``Exclusion'' condition, we enforce the orthogonality constraint through explicit architectural design. Within the outcome regression network (see Section 4.3), we physically block the pathway from $\mathbf{z}_i$ to the outcome prediction. By permitting only the environment $\mathbf{e}_i$ and treatment $t_i$ to influence the outcome, we structurally ensure that the IV affects the outcome only indirectly through the treatment variable.

\subsection{Two-Stage Causal Effect Estimation}
To prevent gradient conflicts and information leakage between the disentangling task and the causal inference task, we adopt a two-stage training strategy.

First, in Stage 1 (i.e., treatment regression), we primarily focus on recovering IVs and fitting the treatment variable. The overall objective function is the weighted sum of multi-task losses:
\begin{equation}
\mathcal{L}_{Stage1} =  \mathcal{L}_{treat} + \beta\mathcal{L}_{ELBO}  + \lambda\mathcal{L}_{ortho},
\end{equation}
This stage focuses on learning a robust latent space where $\mathbf{z}_i$ is effectively decoupled and validated as relevant IVs. Concurrently, treatment variables are fitted to provide treatment inputs for causal effect estimation in the subsequent stage.

Upon convergence of Stage 1, all parameters are frozen to estimate latent outcomes. In Stage 2 (i.e., outcome regression), we employ an independent $\text{GCN}_Y$ to learn the confounder proxy $\mathbf{e}_i$ specifically for outcome regression. This strategy effectively decouples outcome modelling from GCN in Stage 1, which is primarily optimised for treatment fitting, thereby preventing task interference. Subsequently, we construct an outcome regression network $f_Y$ and train it to minimise the Mean Squared Error (MSE):

\begin{equation}
\begin{aligned}
    \mathbf{e}_i^{'} &= \text{GCN}_{Y}(\mathbf{X}, \mathbf{A})_i \\&= \sigma ( \sum_{j \in \mathcal{N}_i \cup \{i\}} \frac{1}{\sqrt{\tilde{d}_i \tilde{d}_j}} \mathbf{W}_{env}^{'} \mathbf{x}_j ),
\end{aligned}
\end{equation}

\begin{equation}
\mathcal{L}_{Stage2} = \frac{1}{N} \sum_{i=1}^N (y_i - f_Y(\mathbf{e}_i^{'}, \hat{t}_i))^2,
\end{equation}

Following prior research~\cite{guo2020learning,zhao2024networked,chen2024doubly}, we utilise the confounder proxy $\mathbf{e}_i$ to correct for confounding bias. The pseudocode for the DisIV algorithm is shown in Algorithm~\ref{alg:disiv}. Additionally, we have analysed the time complexity in the Appendix~\ref{Complexity}. The source code for DisIV is available at \url{https://anonymous.4open.science/r/DisIV}.

\begin{algorithm}[t]
\caption{DisIV Algorithm.}
\label{alg:disiv}
\begin{algorithmic}[1]
\REQUIRE Graph $\mathcal{G}=(\mathbf{X}, \mathbf{A})$, Treatment $T$, Outcome $Y$, Hyperparams $\beta, \lambda$
\ENSURE Trained parameters for Causal Inference

\STATE \textbf{Stage 1: Treatment Regression \& IV Discovery}
\STATE Initialize $\mathbf{W}_{env}, \phi, \theta, \theta_{f_T}$
\WHILE{not converged}
    \STATE $\mathbf{E} \leftarrow \text{GCN}(\mathbf{X}, \mathbf{A}; \mathbf{W}_{env})$ \COMMENT{Eq. (6)}
    \STATE $\mu_\phi, \sigma_\phi \leftarrow \text{Encoder}(\mathbf{X}; \phi)$ \COMMENT{Eq. (7)}
    \STATE $\mathbf{z} \leftarrow \mu_\phi + \sigma_\phi \odot \boldsymbol{\epsilon}$ \COMMENT{Eq. (8)}
    \STATE $\hat{\mathbf{X}} \leftarrow \text{Decoder}(\mathbf{z}, \mathbf{e}; \theta)$ \COMMENT{Eq. (9)}
    \STATE $\mathcal{L}_{ELBO}\leftarrow ELBO(\mathbf{z},\mathbf{e})$ \COMMENT{Eq. (13)}
    \STATE $\mathcal{L}_{ortho}\leftarrow Reg(\mathbf{z},\mathbf{e})$ \COMMENT{Eq. (14)}
    \STATE $\mathcal{L}_{treat} \leftarrow \sigma(f_T([\mathbf{z}, \mathbf{e}]; \theta_{f_T}))$ \COMMENT{Eq. (15)}
    \STATE $\mathcal{L}_{Stage1} \leftarrow \mathcal{L}_{treat} + \beta\mathcal{L}_{ELBO} + \lambda\mathcal{L}_{ortho}$ \COMMENT{Eq. (16)}
    \STATE Update $\{ \mathbf{W}_{env}, \phi, \theta, \theta_{f_T} \}$ minimizing $\mathcal{L}_{Stage1}$
\ENDWHILE

\STATE \textbf{Stage 2: Outcome Regression}
\STATE Freeze $\mathbf{W}_{env}, \phi, \theta, \theta_{f_T}$
\STATE Initialize $\mathbf{W}'_{env}$ and $\theta_{f_Y}$
\WHILE{not converged}
    \STATE $\mathbf{e}' \leftarrow \text{GCN}_Y(\mathbf{X}, \mathbf{A}; \mathbf{W}'_{env})$ \COMMENT{Eq. (17)}
    \STATE $\mathcal{L}_{Stage2} \leftarrow \text{MSE}(Y, f_Y(\mathbf{e}', \hat{t}_i))$ \COMMENT{Eq. (18)}
    \STATE Update $\{ \mathbf{W}'_{env}, \theta_{f_Y} \}$ minimizing $\mathcal{L}_{Stage2}$
\ENDWHILE

\end{algorithmic}
\end{algorithm}

\section{Experiments}
In this section, we conduct extensive experiments on two widely used real-world network datasets to address the following core research questions:
\begin{itemize}
    \item RQ1 (Causal Effect Estimation): Does DisIV outperform existing state-of-the-art baselines in estimating ITE and ATE?
    \item RQ2 (Ablation Study): What contributions do the proposed structural disentangling mechanism and IV validity constraints make to model performance?
    \item RQ3 (Validity of Identification): Does DisIV validly recover the ground-truth IVs and effectively decouple them from environmental confounding?
\end{itemize}

\subsection{Experimental Setup}
% Given the absence of fundamental ground truth for counterfactual outcomes in real-world network observations, we adhere to standard practice within the field by employing a semi-synthetic data generation strategy. We utilise authentic social network topologies $\mathbf{A}$ and user attributes, simulating treatment variables $T$ and outcome variables $Y$ through a rigorously controlled data generation process (DGP).
% In real-world scenarios, observational data lacks counterfactual outcomes. Therefore, following existing research, we generate corresponding data based on authentic datasets.

\textbf{Datasets.} We employed two widely used semi-synthetic datasets, BlogCatalog (BC) and Flickr~\cite{jiang2022estimating}, to validate the effectiveness of the proposed DisIV method. Due to space constraints, detailed information about the datasets is provided in the Appendix~\ref{appendix_Datasets}.

\begin{table*}[t]
    \centering
    \caption{\textbf{Results on BC Dataset.} We report the Mean $\pm$ Std of $\sqrt{\epsilon_{PEHE}}$ and $\epsilon_{ATE}$. Here, ``a-b'' denotes the weighting configuration with $\mathbf{w}_{C}=a$ and $\mathbf{w}_{U}=b$ regulating the intensity of $C_{net}$ and $U$, respectively. The best results are highlighted in \textbf{bold}.}
    \label{BC}
\begin{center}
\begin{small}
  \begin{sc}
    \begin{tabular}{lcccccc}
        \toprule
        \multirow{2}{*}{\textbf{Method}} & \multicolumn{3}{c}{\textbf{Within-sample}} & \multicolumn{3}{c}{\textbf{Out-of-sample}} \\
        \cmidrule(lr){2-4} \cmidrule(lr){5-7}
        & \textbf{0.5-0.5} & \textbf{0.5-1.0} & \textbf{1.0-1.0} & \textbf{0.5-0.5} & \textbf{0.5-1.0} & \textbf{1.0-1.0} \\
        \midrule
        
        % --- Metric: PEHE ---
        \multicolumn{7}{c}{\textit{Metric: $\sqrt{\epsilon_{PEHE}}\downarrow$}} \\
        \midrule
        CFR             & 0.78 $\pm$ 0.43 & 0.97 $\pm$ 0.65 & 1.35 $\pm$ 0.91 & 0.78 $\pm$ 0.44 & 1.34 $\pm$ 0.69 & 1.35 $\pm$ 0.91 \\
        TARNET          & 1.42 $\pm$ 0.50 & 2.11 $\pm$ 0.74 & 2.01 $\pm$ 1.57 & 1.46 $\pm$ 0.54 & 2.18 $\pm$ 0.80 & 2.01 $\pm$ 1.56 \\
        NetDeconf       & 0.88 $\pm$ 0.52 & 1.62 $\pm$ 0.89 & 1.55 $\pm$ 0.19 & 0.83 $\pm$ 0.43 & 1.55 $\pm$ 0.91 & 1.34 $\pm$ 0.25 \\
        DeepIV          & 1.31 $\pm$ 1.88 & 1.23 $\pm$ 0.86 & 3.27 $\pm$ 2.77 & 1.31 $\pm$ 1.88 & 1.22 $\pm$ 0.86 & 3.27 $\pm$ 2.77 \\
        NetIV           & 0.97 $\pm$ 0.63 & 1.49 $\pm$ 0.62 & 1.48 $\pm$ 0.24 & 0.91 $\pm$ 0.65 & 1.33 $\pm$ 0.59 & 1.31 $\pm$ 0.29 \\
        \textbf{DisIV} & \textbf{0.73 $\pm$ 0.35} & \textbf{0.77 $\pm$ 0.25} & \textbf{1.09 $\pm$ 0.36} & \textbf{0.69 $\pm$ 0.34} & \textbf{0.72 $\pm$ 0.24} & \textbf{1.13 $\pm$ 0.39} \\
        
        \midrule
        % --- Metric: ATE ---
        \multicolumn{7}{c}{\textit{Metric: $\epsilon_{ATE}\downarrow$}} \\
        \midrule
        CFR             & 0.64 $\pm$ 0.33 & 1.33 $\pm$ 0.71 & 1.34 $\pm$ 0.92 & 0.62 $\pm$ 0.33 & 1.33 $\pm$ 0.70 & 1.34 $\pm$ 0.92 \\
        TARNET          & 0.83 $\pm$ 0.57 & 1.60 $\pm$ 0.69 & 1.95 $\pm$ 1.63 & 0.88 $\pm$ 0.52 & 1.67 $\pm$ 0.58 & 1.95 $\pm$ 1.63 \\
        NetDeconf       & 0.41 $\pm$ 0.32 & 1.21 $\pm$ 0.99 & 0.82 $\pm$ 0.48 & 0.47 $\pm$ 0.36 & 1.18 $\pm$ 1.00 & 0.70 $\pm$ 0.52 \\
        DeepIV          & 1.26 $\pm$ 1.92 & 1.21 $\pm$ 0.87 & 3.26 $\pm$ 2.77 & 1.26 $\pm$ 1.92 & 1.21 $\pm$ 0.87 & 3.26 $\pm$ 2.78 \\
        NetIV           & 0.60 $\pm$ 0.68 & 0.52 $\pm$ 0.48 & 0.65 $\pm$ 0.52 & 0.60 $\pm$ 0.74 & 0.52 $\pm$ 0.45 & 0.60 $\pm$ 0.48 \\
        \textbf{DisIV} & \textbf{0.40 $\pm$ 0.43} & \textbf{0.41 $\pm$ 0.27} & \textbf{0.57 $\pm$ 0.58} & \textbf{0.40 $\pm$ 0.43} & \textbf{0.40 $\pm$ 0.27} & \textbf{0.60 $\pm$ 0.59} \\
        \bottomrule
    \end{tabular}
  \end{sc}
\end{small}
\end{center}
    \vskip -0.1in
\end{table*}

% =========================================================================
% Table 2: Flickr Dataset Results
% =========================================================================
\begin{table*}[t]
    \centering
    \caption{\textbf{Results on Flickr Dataset.} We report the Mean $\pm$ Std of $\sqrt{\epsilon_{PEHE}}$ and $\epsilon_{ATE}$. The best results are highlighted in \textbf{bold}.}
    \label{Flickr}
\begin{center}
\begin{small}
  \begin{sc}
    \begin{tabular}{lcccccc}
        \toprule
        \multirow{2}{*}{\textbf{Method}} & \multicolumn{3}{c}{\textbf{Within-sample}} & \multicolumn{3}{c}{\textbf{Out-of-sample}} \\
        \cmidrule(lr){2-4} \cmidrule(lr){5-7}
        & \textbf{0.5-0.5} & \textbf{0.5-1.0} & \textbf{1.0-1.0} & \textbf{0.5-0.5} & \textbf{0.5-1.0} & \textbf{1.0-1.0} \\
        \midrule
        
        % --- Metric: PEHE ---
        \multicolumn{7}{c}{\textit{Metric: $\sqrt{\epsilon_{PEHE}}\downarrow$}} \\
        \midrule
        CFR             & 2.50 $\pm$ 1.41 & 1.69 $\pm$ 1.49 & 1.32 $\pm$ 1.15 & 2.50 $\pm$ 1.41 & 1.69 $\pm$ 1.49 & 1.33 $\pm$ 1.15 \\
        TARNET          & 2.82 $\pm$ 1.42 & 2.38 $\pm$ 0.72 & 4.85 $\pm$ 5.02 & 2.79 $\pm$ 1.41 & 2.38 $\pm$ 0.72 & 4.81 $\pm$ 5.04 \\
        NetDeconf       & 1.55 $\pm$ 0.41 & 1.77 $\pm$ 0.70 & 1.83 $\pm$ 0.44 & 1.29 $\pm$ 0.39 & 1.67 $\pm$ 0.46 & 1.73 $\pm$ 0.63 \\
        DeepIV          & 3.20 $\pm$ 2.45 & 1.26 $\pm$ 0.45 & 3.08 $\pm$ 3.34 & 3.21 $\pm$ 2.45 & 1.26 $\pm$ 0.45 & 3.08 $\pm$ 3.33 \\
        NetIV           & 1.86 $\pm$ 0.65 & 1.93 $\pm$ 0.78 & 2.35 $\pm$ 0.92 & 1.30 $\pm$ 0.57 & 1.52 $\pm$ 0.52 & 1.74 $\pm$ 0.47 \\
        \textbf{DisIV} & \textbf{0.85 $\pm$ 0.39} & \textbf{0.83 $\pm$ 0.21} & \textbf{1.13 $\pm$ 0.16} & \textbf{0.82 $\pm$ 0.37} & \textbf{0.89 $\pm$ 0.16} & \textbf{1.08 $\pm$ 0.27} \\
        
        \midrule
        % --- Metric: ATE ---
        \multicolumn{7}{c}{\textit{Metric: $\epsilon_{ATE}\downarrow$}} \\
        \midrule
        CFR             & 2.49 $\pm$ 1.42 & 1.65 $\pm$ 1.53 & 1.30 $\pm$ 1.16 & 2.49 $\pm$ 1.41 & 1.65 $\pm$ 1.52 & 1.31 $\pm$ 1.16 \\
        TARNET          & 1.70 $\pm$ 1.50 & 2.33 $\pm$ 0.75 & 3.46 $\pm$ 5.37 & 1.65 $\pm$ 1.55 & 2.33 $\pm$ 0.75 & 3.46 $\pm$ 5.41 \\
        NetDeconf       & 0.85 $\pm$ 0.49 & 0.79 $\pm$ 0.70 & 0.86 $\pm$ 0.59 & 0.73 $\pm$ 0.47 & 0.82 $\pm$ 0.54 & 0.83 $\pm$ 0.61 \\
        DeepIV          & 3.20 $\pm$ 2.45 & 1.25 $\pm$ 0.46 & 3.07 $\pm$ 3.34 & 3.20 $\pm$ 2.46 & 1.25 $\pm$ 0.46 & 3.07 $\pm$ 3.34 \\
        NetIV           & 0.71 $\pm$ 0.37 & 0.89 $\pm$ 0.72 & 0.97 $\pm$ 0.62 & 0.67 $\pm$ 0.38 & 0.89 $\pm$ 0.54 & 0.95 $\pm$ 0.51 \\
        \textbf{DisIV} & \textbf{0.66 $\pm$ 0.45} & \textbf{0.37 $\pm$ 0.39} & \textbf{0.41 $\pm$ 0.29} & \textbf{0.63 $\pm$ 0.44} & \textbf{0.39 $\pm$ 0.41} & \textbf{0.42 $\pm$ 0.23} \\
        \bottomrule
    \end{tabular}
  \end{sc}
\end{small}
\end{center}
    \vskip -0.1in
\end{table*}

\textbf{Data Generation Process.} Due to the absence of counterfactual outcomes in real-world data, we constructed semi-synthetic data incorporating feature entanglement properties. First, we decoupled observed features into latent IVs $\mathbf{Z}_{true}$ and environmental confounders $\mathbf{C}_{net}$, while introducing unobservable confounders $\mathbf{U}$. When generating the treatment variable $t_i$ and outcome variable $y_i$, we introduced key weighting parameters $\mathbf{w}_{X}$, $\mathbf{w}_{IV}$, $\mathbf{w}_{C}$, and $\mathbf{w}_{U}$, which control the weights of different variables respectively. It is noteworthy that the generation of the outcome variable $y_i$ strictly excludes $\mathbf{Z}_{true}$ to satisfy the IV's exclusivity assumption. Detailed generation formulas and parameter settings are provided in the Appendix~\ref{DGP}.

\textbf{Validation Metrics.} We adopt two standard metrics. The Precision in Estimation of Heterogeneous Effect ($\sqrt{\epsilon_{PEHE}}$) measures the root mean squared error of ITE estimates, defined as:
\begin{equation}
    \sqrt{\epsilon_{PEHE}} = \sqrt{\frac{1}{N} \sum_{i=1}^N (\hat{\tau}_i - \tau_i)^2}.
\end{equation}
The ATE error ($\epsilon_{ATE}$) quantifies the absolute bias of the ATE, defined as:
\begin{equation}
   \epsilon_{ATE} = | \hat{\tau}_{ATE} - \tau_{ATE} |.
\end{equation}

\textbf{Baseline.} To comprehensively evaluate performance, we compare DisIV with five representative baselines spanning classical causal inference and network-aware deep learning methods. \textit{CFR} mitigates confounding via IPM-based distribution balancing but ignores network structure, while \textit{TARNet} removes the balancing regularisation and relies solely on representation learning. \textit{NetDeconf} leverages GCNs for confounder adjustment, assuming network information can proxy latent confounders. \textit{DeepIV} serves as a general deep IV baseline; in the absence of true IVs, we use $\mathbf{X}$ together with the $\mathbf{A}$ as IV inputs. \textit{NetIV} is a recent method for recovering IVs from network observation data. Detailed baseline information is provided in the Appendix~\ref{Baseline}.

Due to space constraints, we present the \textbf{Experimental Details} in the \ref{appendix_Details}.

\begin{table}[t]
    \centering
    \caption{\textbf{Ablation Study Results.} We report the Mean $\pm$ Std of $\sqrt{\epsilon_{PEHE}}$ and $\epsilon_{ATE}$ on the BC dataset.}
    \label{tab:ablation_metric_wise}
    \begin{center}
    \begin{small}
    \begin{sc}
        \begin{tabular}{lcc}
            \toprule
            \multirow{2}{*}{\textbf{Method}} & \textbf{Within-sample} & \textbf{Out-of-sample} \\
            \cmidrule(lr){2-2} \cmidrule(lr){3-3}
             & \textbf{0.5-0.5} & \textbf{0.5-0.5} \\
            \midrule
            
            % --- Metric: PEHE ---
            \multicolumn{3}{c}{\textit{Metric: $\sqrt{\epsilon_{PEHE}}\downarrow$}} \\
            \midrule
            DisIV-VAE   & 0.80 $\pm$ 0.64 & 0.84 $\pm$ 0.64 \\
            DisIV-w/o Reg  & 0.81 $\pm$ 0.60 & 0.80 $\pm$ 0.59 \\
            \textbf{DisIV} & \textbf{0.73 $\pm$ 0.35} & \textbf{0.69 $\pm$ 0.34} \\
            
            \midrule
            % --- Metric: ATE ---
            \multicolumn{3}{c}{\textit{Metric: $\epsilon_{ATE}\downarrow$}} \\
            \midrule
            DisIV-VAE   & 0.58 $\pm$ 0.75 & 0.59 $\pm$ 0.76 \\
            DisIV-w/o Reg  & 0.58 $\pm$ 0.70 & 0.58 $\pm$ 0.68 \\
            \textbf{DisIV} & \textbf{0.40 $\pm$ 0.43} & \textbf{0.40 $\pm$ 0.43} \\
            \bottomrule
        \end{tabular}
    \end{sc}
    \end{small}
    \end{center}
    \vskip -0.1in
\end{table}
\subsection{Results Comparison (RQ1)}
We systematically compared DisIV against five state-of-the-art (SOTA) baseline methods across two datasets (BC and Flickr), comprehensively evaluating both the ITE and ATE. Experiments spanned both within-sample and out-of-sample settings, examining model performance under varying levels of confounding intensity.

As demonstrated in Tables~\ref{BC} and~\ref{Flickr}, DisIV achieved optimal performance across all experimental settings. Firstly, compared to IID-based methods (such as CFR and TARNet) that neglect data dependencies, DisIV's explicit modelling of neighbour information enables more precise adjustment for network-induced confounding biases. This advantage is particularly pronounced in the larger, structurally more complex Flickr dataset. Secondly, although NetDeconf also utilises neighbour information, DisIV further explicitly decouples individual-specific variation in observational data from environmental proxies, thereby enabling finer-grained bias removal. Finally, compared with IV-based methods, DisIV effectively mitigates DeepIV's reliance on predefined IVs while avoiding the confounding noise introduced by NetIV's coarse treatment of neighbours as IVs. By constructing pure latent IVs, DisIV ensures the validity of causal inference.

\subsection{Ablation Study (RQ2)}
To validate the efficacy of each component within DisIV, we conducted ablation experiments and constructed the following two variants: (1) \textit{DisIV-VAE}. This variant removes the conditional constraint on the confounder proxy $\mathbf{z}_i$ in Decoder $p_\theta(\mathbf{x}_i|\mathbf{z}_i, \mathbf{e}_i)$. (2) \textit{DisIV-w/o Reg}. This variant removes the orthogonality regularisation term.

As shown in Table~\ref{tab:ablation_metric_wise}, DisIV demonstrates significantly superior performance compared to its variants, confirming the effectiveness of its key components. Specifically, the performance degradation of DisIV-VAE indicates that the absence of decoder conditional constraints forces environmental information to leak into $\mathbf{z}$. Meanwhile, the inferiority of DisIV-w/o Reg demonstrates that structure alone is insufficient to fully eliminate residual correlations. This suggests that the asymmetric structure provides the disentangling foundation, while orthogonal regularisation offers stringent safeguards; their synergy ensures the purity of the latent IVs. More experimental results are presented in the Appendix~\ref{Ablation_Flickr}.

% As shown in Table~\ref{tab:ablation_metric_wise}, both Dis-VAE and DisIV-w/o Reg exhibited significantly inferior performance compared to DisIV. Firstly, DisIV-VAE exhibits a marked performance decline. This indicates that without the confounder proxy as a decoder input, the encoder is compelled to embed environmental confounding information into the latent IV $\mathbf{z}$ to minimise reconstruction error. Secondly, DisIV-w/o Reg's performance degradation demonstrates the necessity of explicit constraints. Although the asymmetric structure tends to promote separation, the absence of orthogonal regularisation makes it difficult for the model to completely eliminate residual correlations between $\mathbf{z}$ and the environment, thereby introducing estimation bias. In summary, the asymmetric structure provides the foundational framework for disentangling, while orthogonal regularisation offers rigorous mathematical guarantees. Together, they ensure that DisIV can extract valid latent IVs from observational data.

\subsection{Analysis of Latent Disentanglement (RQ3)}
To validate the effectiveness of DisIV, we employed the coefficient of determination ($R^2$ Score) to quantitatively assess the linear alignment between the learned latent representations ($\mathbf{z}, \mathbf{u}$) and the ground-truth (GT) factors.

% To assess the validity of the recovered latent variables, we used the coefficient of determination ($R^2$ Score) to quantify the alignment between the representations learned by DisIV and the true factors during data synthesis. Specifically, we computed linear fits between $\mathbf{z}$ and $\mathbf{u}$, and between the ground-truth IV and the ground-truth confounding.

\begin{figure}
    \centering
    \includegraphics[width=0.78\columnwidth]{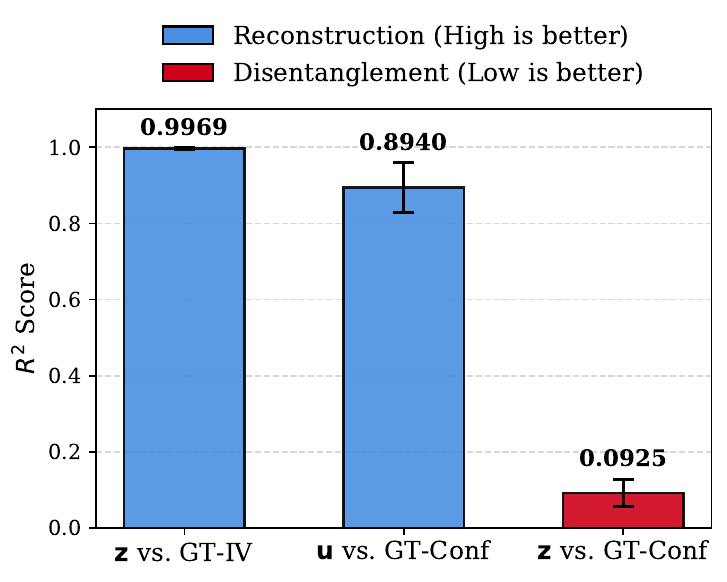}
    \caption{Quantitative analysis of latent factor validity on the BC dataset. The GT-IV denotes $\mathbf{Z}_{true}$, and the GT-Conf denotes $\mathbf{C}_{net}$.}
    \label{Disentanglement}
\end{figure}

As shown in Figure~\ref{Disentanglement}, the latent factors recovered by DisIV exhibit an exceptionally high degree of alignment with the GT factors (high $R^2$). Crucially, the $\mathbf{z}$ maintains an extremely low $R^2$ level with the true environmental confounders. This result conclusively demonstrates that DisIV successfully decouples the latent IVs from confounding factors while capturing useful information from the observed data. More experimental results are presented in the Appendix~\ref{Disentanglement_Flickr}.

% As demonstrated in the results, the DisIV-recovered IV $\mathbf{z}$ exhibits an exceptionally high $R^2$ score relative to the true IV. Concurrently, the confounding proxy captured by DisIV also demonstrates a strong correlation with the true environmental confounding, indicating the model accurately reconstructs the underlying generative factors. Crucially, the $R^2$ score between $\mathbf{z}$ and the true environmental confounding remains at an extremely low level. In summary, DisIV not only successfully captures relevant information from observed data but also effectively decouples the latent IV from confounding factors, fully demonstrating the method's validity.

\section{Conclusion}
In this paper, we propose DisIV, a disentangled IV framework for networked environments that explicitly separates individual-specific information (i.e., latent IVs) from confounder information in observational data. Specifically, we devise an asymmetric inference-generation structure to extract latent factors, combined with a constrained optimisation mechanism to synergistically purify IVs. Extensive experimental results demonstrate that DisIV effectively decouples high-quality latent IVs, thereby enabling unbiased causal effect estimation within complex network settings.

\section*{Impact Statement}
This paper presents work whose goal is to advance the field of Machine Learning. There are many potential societal consequences of our work, none which we feel must be specifically highlighted here.
% Note use of \abovespace and \belowspace to get reasonable spacing
% above and below tabular lines.


\bibliography{ref}
\bibliographystyle{icml2026}

%%%%%%%%%%%%%%%%%%%%%%%%%%%%%%%%%%%%%%%%%%%%%%%%%%%%%%%%%%%%%%%%%%%%%%%%%%%%%%%
%%%%%%%%%%%%%%%%%%%%%%%%%%%%%%%%%%%%%%%%%%%%%%%%%%%%%%%%%%%%%%%%%%%%%%%%%%%%%%%
% APPENDIX
%%%%%%%%%%%%%%%%%%%%%%%%%%%%%%%%%%%%%%%%%%%%%%%%%%%%%%%%%%%%%%%%%%%%%%%%%%%%%%%
%%%%%%%%%%%%%%%%%%%%%%%%%%%%%%%%%%%%%%%%%%%%%%%%%%%%%%%%%%%%%%%%%%%%%%%%%%%%%%%
\newpage
\appendix
\onecolumn
\section{Instrumental Variable}
\label{appendix_IV}
Instrumental Variables (IVs) are exogenous variables that influence outcomes solely through their strong association with the treatment variable. The valid IVs should satisfy the following three assumptions:
\begin{definition}[Valid Instrumental Variable]
    A latent variable $\mathbf{z}_i$ constitutes a valid IV if and only if it satisfies:
    \begin{itemize}
        \item Relevance: $\mathbf{z}_i \not\perp t_i$. The latent variable $\mathbf{z}_i$ must be correlated with the treatment variable $t_i$.
        \item Exclusion: $P(y_i \mid t_i, \mathbf{z}_i, \mathbf{u}_i) = P(y_i \mid t_i, \mathbf{u}_i)$. That is, $\mathbf{z}_i$ has no direct effect on the outcome, exerting its influence solely indirectly through the treatment variable.
        \item Unconfoundedness: $\mathbf{z}_i \perp \mathbf{u}_i$. The latent variable $\mathbf{z}_i$ must remain independent of the confounders $\mathbf{u}_i$.
    \end{itemize}
\end{definition}

\section{Complexity Analysis}
\label{Complexity}
We provide a detailed complexity analysis of DisIV. Let $N$ and $M$ denote the number of nodes and edges, $D$ the input dimension, and $d$ the latent dimension. In our implementation, $d$ is significantly larger than the reduced input dimension $D$.

The computational cost per epoch is determined by two key components:
\begin{itemize}
    \item \textbf{Graph Aggregation (GCN):} The environment encoder employs $L$ GCN layers. Each layer performs sparse matrix multiplication for neighbour aggregation, scaling as $\mathcal{O}(L \cdot M \cdot d)$.
    \item \textbf{Asymmetric Inference \& Generation:} This part involves dense matrix operations. The \textit{Encoder} maps input features to latent factors ($\mathbb{R}^D \to \mathbb{R}^d$), costing $\mathcal{O}(N \cdot D \cdot d)$. The \textit{Decoder} reconstructs features conditioned on the proxy ($\mathbb{R}^{2d} \to \mathbb{R}^D$), scaling as $\mathcal{O}(N \cdot 2d \cdot D)$. Crucially, since $d \gg D$, the dominant cost arises from the internal transformations within the hidden layers of the MLPs, which scales as $\mathcal{O}(N \cdot d^2)$.
\end{itemize}

Since $L$, $D$, and $d$ are constants relative to the graph size, the complexity remains linear with respect to the number of nodes and edges, i.e., $\mathcal{O}(N + M)$. This ensures that DisIV is highly efficient and scalable to large-scale networked data. Summing these components over $E$ epochs, the total time complexity is:
\begin{equation}
    \mathcal{T}_{\text{total}} \approx \mathcal{O}\big( E \cdot (L M d + N d^2) \big).
\end{equation}

\section{Datasets}
\label{appendix_Datasets}
We employed two widely used semi-synthetic datasets, BlogCatalog and Flickr, to validate the effectiveness of the proposed DisIV method.
\begin{itemize}
    \item \textbf{BlogCatalog (BC)} is a social network dataset derived from a blogging platform, where nodes represent users and edges denote social links. Node attributes are constructed from keywords extracted from blog content.
    \item \textbf{Flickr} is a user interaction network with follow relationships, where node features are constructed from image tags reflecting users’ content preferences.
\end{itemize}
The statistical information for the BC and Flickr datasets is shown in Table~\ref{datasets}.
\begin{table}[h]
\centering
\caption{Statistics of the datasets.}
\label{datasets}
\begin{tabular}{lcc}
\toprule
Dataset & \# Nodes & \# Edges \\
\midrule
BlogCatalog & 5,196 & 171,743 \\
Flickr      & 7,575 & 239,738 \\
\bottomrule
\end{tabular}
\vskip -0.1in
\end{table}

% \section{Validation Metrics.}
% \label{appendix_Metrics}
% We adopt two standard metrics. The Precision in Estimation of Heterogeneous Effect ($\sqrt{\epsilon_{PEHE}}$) measures the root mean squared error of ITE estimates, defined as:
% \begin{equation}
%     \sqrt{\epsilon_{PEHE}} = \sqrt{\frac{1}{N} \sum_{i=1}^N (\hat{\tau}_i - \tau_i)^2}.
% \end{equation}
% The ATE error ($\epsilon_{ATE}$) quantifies the absolute bias of the ATE, defined as:
% \begin{equation}
%    \epsilon_{ATE} = | \hat{\tau}_{ATE} - \tau_{ATE} |.
% \end{equation}

\section{Data Generation Process}
\label{DGP}
In real-world scenarios, observational data lacks counterfactual outcomes. Therefore, following existing research~\cite{guo2020learning,zhao2024networked,jiang2022estimating}, we generate corresponding data based on real datasets. To address high-dimensional sparse attributes, we first employ Latent Dirichlet Allocation (LDA) to reduce the dimensionality of raw node attributes to $K=10$ dimensions, thereby obtaining a high-density feature matrix $\mathbf{X}_{raw}$. To simulate feature entanglement, we devise the following generation mechanism:

First, we partition the reduced-dimension feature matrix $\mathbf{X}_{raw}$ along the feature dimension into two subspaces:
\begin{equation}
    \mathbf{X}_{raw} = [\mathbf{X}_{IV} \| \mathbf{X}_{Conf}],
\end{equation}
simulating latent IV sources and environmental confounding sources, respectively. The generation mechanisms for $\mathbf{X}_{IV}$ and $\mathbf{X}_{Conf}$ are as follows:
\begin{itemize}
    \item Construction of Latent IVs: To model implicit and highly entangled IV in real-world scenarios, we design latent IVs as hidden variables generated from individual-specific features via multi-layer nonlinear mappings. Specifically, we apply the following transformation to $\mathbf{X}_{IV}$: 
    \begin{equation}
        \mathbf{Z}_{true} = \tanh(\mathbf{X}_{IV} \mathbf{W}_{proj}), 
    \end{equation}
    where $\mathbf{W}_{proj}$ is a random projection matrix. This ensures a complex nonlinear relationship between the IV and the original features.
    \item Construction of environmental confounding: We introduce network homogeneity effects to the latter half of the features:
    \begin{equation}
        \mathbf{C}_{net} = \mathbf{A} \mathbf{X}_{Conf}.
    \end{equation}
    This simulates the characteristic of environmental confounding, often defined by neighbour structures.
\end{itemize}
Beyond the observed features, we introduce an unobserved confounder $\mathbf{U} \sim \mathcal{N}(0, 1)$ to model confounding factors that are inherently unobservable in practice.

Subsequently, we simulate the assignment of intervention strategies. The binary treatment variable $t_i \in \{0, 1\}$ is jointly determined by the individual's specific IV, environmental confounders, raw features, and unobserved confounders. The generation formula is as follows:
\begin{equation}
    t_i \sim \text{Bernoulli}(\sigma(\mathcal{L}_i)),
\end{equation}
\begin{equation}
    \mathcal{L}_i = \mathbf{Z}_{true} \mathbf{w}_{IV} + \mathbf{C}_{net} \mathbf{w}_{C} + \mathbf{X}_{raw} \mathbf{w}_{X} + \mathbf{U} \mathbf{w}_{U}.
\end{equation}
where $\sigma(\cdot)$ denotes the Sigmoid function, and each weight coefficient controls the strength of influence exerted by different factors on treatment assignment. This construction introduces strong endogeneity while ensuring $\mathbf{Z}_{true}$ remains significantly correlated with the treatment variable, thereby satisfying the correlation assumption at the data level.

Finally, we simulate the generation of the Outcome. The outcome variable $y_i$ is generated as follows:
\begin{equation}
    y_i = \beta_T t_i + \mathbf{C}_{net} \mathbf{w}_{C} + \mathbf{X}_{raw} \mathbf{w}_{X} + \mathbf{U} \mathbf{w}_{U} + \epsilon,
\end{equation}
where $\epsilon$ is an IID noise term. It is particularly noteworthy that $\mathbf{Z}_{true}$ is explicitly excluded from the outcome generation formula. Its influence on the outcome variable is entirely mediated through the treatment variable $t_i$, thereby strictly satisfying the exclusivity constraint at the data generation level. Should the model accurately recover $\mathbf{Z}_{true}$ and employ it as an IV, it would eliminate the confounding bias arising from $\mathbf{U}$ and $\mathbf{C}_{net}$.

\section{Baseline}
\label{Baseline}
To comprehensively evaluate performance, we compare DisIV with five representative baselines spanning classical causal inference and network-aware deep learning methods:
\begin{itemize}
    \item \textit{Treatment-Agnostic Representation Network (TARNet)~\cite{shalit2017estimating}:} TARNet employs a dual-head architecture to learn representations, eschewing the treatment of interventions as mere covariates. TARNet projects input features into a shared latent space, which then bifurcates into independent branches to model the potential outcomes for treated and control groups separately.
    
    \item \textit{Counterfactual Regression (CFR)~\cite{shalit2017estimating}:} Building upon TARNet, CFR integrates a representation balancing mechanism to mitigate selection bias. By imposing an Integral Probability Metric (IPM, e.g., Wasserstein distance) regularization, it minimizes the distributional discrepancy between treated and control groups within the latent feature space.
    
    \item \textit{Network Deconfounder (NetDeconf)~\cite{guo2020learning}:} NetDeconf addresses unobserved confounders in networked data by leveraging the network homophily assumption. NetDeconf utilizes Graph Convolutional Networks (GCNs) to aggregate neighbour information into a confounder proxy, which conditions the outcome regression to rectify network-induced confounding bias.
    
    \item \textit{DeepIV~\cite{hartford2017deep}:} DeepIV extends the classical IV framework to deep neural architectures for handling unobserved confounders in nonlinear settings. DeepIV adopts a two-stage procedure: estimating the conditional treatment distribution, followed by optimizing a weighted loss function to fit the outcome model.
    
    \item \textit{Networked IV (NetIV)~\cite{zhao2024networked}:} NetIV exploits network topology to construct endogenous instruments in the absence of external IVs. Capitalizing on peer influence, NetIV employs Graph Neural Networks (GNNs) to extract IV signals from neighbour information.
\end{itemize}

\section{Experimental Details}
\label{appendix_Details}
All models were implemented using PyTorch and trained on an NVIDIA RTX 4090 GPU (24GB). Hyperparameters for all methods were selected via grid search on the validation set, with hidden layer dimensions set to $128$ and the Adam optimiser employed. To mitigate randomness, experiments were conducted across five predefined data splits, with results presented as the mean $\pm$ standard deviation across multiple runs. During the generation of the semi-synthetic dataset, we set $\mathbf{w}_{C}$ and $\mathbf{w}_{U}$ to $\{0.5, 1.0\}$, with other parameters set to $1.0$.

\section{Ablation Study on the Flickr Dataset}
\label{Ablation_Flickr}
The Table~\ref{tab:ablation_metric_Flickr} presents a performance comparison of DisIV and its variants on the Flickr dataset. The results show that the complete model DisIV achieves optimal performance across most metrics, particularly exhibiting a significant advantage in $\sqrt{\epsilon_{PEHE}}$, which reflects the accuracy of individual heterogeneity estimation.
In comparison, DisIV-VAE, which removes the conditional decoding structure, performs the worst, confirming the crucial role of asymmetric structures in preventing information leakage. The performance degradation of DisIV-w/o Reg on $\sqrt{\epsilon_{PEHE}}$, which removes the orthogonality constraint, demonstrates the necessity of explicit independence constraints for stripping residual confounding and ensuring the purity of IVs. In summary, both key components are indispensable for achieving effective latent IV discovery.
\begin{table}[t]
    \centering
    \caption{\textbf{Ablation Study Results.} We report the Mean $\pm$ Std of $\sqrt{\epsilon_{PEHE}}$ and $\epsilon_{ATE}$ on the Flickr dataset.}
    \label{tab:ablation_metric_Flickr}
    \begin{center}
    \begin{small}
    \begin{sc}
        \begin{tabular}{lcc}
            \toprule
            \multirow{2}{*}{\textbf{Method}} & \textbf{Within-sample} & \textbf{Out-of-sample} \\
            \cmidrule(lr){2-2} \cmidrule(lr){3-3}
             & \textbf{0.5-0.5} & \textbf{0.5-0.5} \\
            \midrule
            
            % --- Metric: PEHE ---
            \multicolumn{3}{c}{\textit{Metric: $\sqrt{\epsilon_{PEHE}}\downarrow$}} \\
            \midrule
            DisIV-VAE   & 1.08 $\pm$ 0.49 & 1.10 $\pm$ 0.54 \\
            DisIV-w/o Reg  & 1.08 $\pm$ 0.35 & 1.05 $\pm$ 0.41 \\
            \textbf{DisIV} & \textbf{0.85 $\pm$ 0.39} & \textbf{0.82 $\pm$ 0.37} \\
            
            \midrule
            % --- Metric: ATE ---
            \multicolumn{3}{c}{\textit{Metric: $\epsilon_{ATE}\downarrow$}} \\
            \midrule
            DisIV-VAE   & 0.72 $\pm$ 0.63 & 0.73 $\pm$ 0.65 \\
            DisIV-w/o Reg  & \textbf{0.65 $\pm$ 0.56} & 0.63 $\pm$ 0.58 \\
            \textbf{DisIV} & 0.69 $\pm$ 0.34 & \textbf{0.63 $\pm$ 0.44} \\
            \bottomrule
        \end{tabular}
    \end{sc}
    \end{small}
    \end{center}
    \vskip -0.1in
\end{table}

\section{Analysis of Latent Disentanglement on the Flickr Dataset}
\label{Disentanglement_Flickr}
\begin{figure}
    \centering
    \includegraphics[width=0.5\columnwidth]{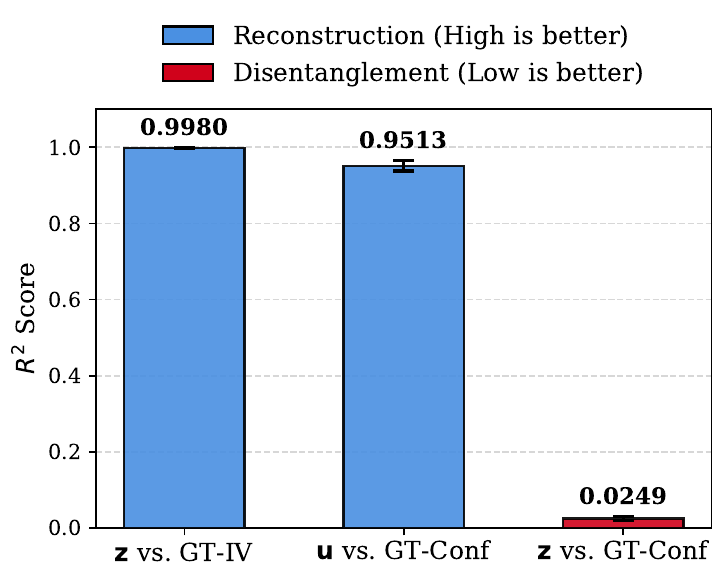}
    \caption{Quantitative analysis of latent factor validity on the Flickr dataset. The GT-IV denotes $\mathbf{Z}_{true}$, and the GT-Conf denotes $\mathbf{C}_{net}$.}
    \label{Disentanglement_Flickr_result}
\end{figure}

Our experiments on the Flickr dataset further verify the decoupling ability of DisIV, as shown in Figure~\ref{Disentanglement_Flickr_result}. The results show that the latent factors recovered by DisIV accurately reflect the true factors, with a high corresponding $R^2$ value; while the $R^2$ value between the latent IV $\mathbf{z}$ and the true confounder is close to zero, indicating that $\mathbf{z}$ successfully captures individual-specific information without being affected by environmental confounders. The results in the Flickr dataset are consistent with those in the BC dataset, verifying its ability to effectively distinguish between latent IV and confounder under different data structures.

\end{document}